# Process mining classification with a weightless neural network

Rafael Garcia Barbastefano[1[0000-0001-8253-6308]], Maria Clara Lippi[1[0000-0002-3540-1301]], and Diego Carvalho[1[0000-0003-1592-6327]]

[1] Federal Centre for Engineering Studies and Technological Education - CEFET/RJ, Rio de Janeiro/RJ, Brazil
d.carvalho@ieee.org

**Abstract.** Using a weightless neural network architecture WiSARD we propose a straightforward graph to retina codification to represent business process graph flows avoiding kernels, and we present how WiSARD outperforms the classification performance with small training sets in the process mining context.

## 1 Introduction

Information systems, as a relevant part of organizations, are strongly linked with business process models and the operations they represent and support [1]. They generate and record data based on event logs, which are especially useful for process mining proposes [2].

Process mining is a research area which merges Business Process Management approaches with data mining tasks and techniques. De Leoni, van der Aalst [3] consider it as “the missing link between model-based process analysis and data-oriented analysis techniques”. Process mining is covered by three types: process discovery, process conformance and process enhancement [4]. Process discovery consists on modeling business processes, as they really occurred, from raw data (majorly system's event logs). The conformance type of process mining consists of comparisons between an existing process and a model [5]. Conformance checking is essential for business, to find undesirable deviations or to measure efficiency.

The literature has several references on the relevance of neural networks in process mining [6], both in process discovery studies and conformance analysis. Maita et al. [7] conducted a systematic review about techniques employed on process mining investigations. They evidenced that computational intelligence or machine learning techniques were present in 19% of the cases, whereas traditional data mining ones represent 81%. Their study found there are no studies combining conformance checking with artificial neural networks and the predominant method is the support vector machines (SVM). SVM cases make use of kernel methods introducing a preprocessing phase when working with graphs. Besides, it suggests the use of methods that need the construction of representative kernels of graph data.

Considering the application of neural networks in mining processes, it is always desirable to have both a clear preprocessing step as well as the use of a small subset of training data. Therefore, it is valuable to measure and compare not only the results but also the methods' performance. Learning curves (LC) are renowned for accomplishing this purpose since it enables to measure predictive performance for different learning effort levels [8]. A classic application example of this tool is the predictive efficiency as a function of training sample size [8].

We propose a graph-to-retina's codification to represent graph flows avoiding kernels, and we present how WiSARD performs on graph flow classification with different levels of dataset complexity visualized by LC. The rest of the paper is organized as follows. A short presentation of the WiSARD weightless neural network architecture is described at Section 2. The process dataset details are given at Section 3 and the experiment design and results in Section 4. Section 5 concludes this paper.

## 2 The WiSARD weightless neural network architecture

The WiSARD weightless neural network name comes from its authors (Wilkie, Stonham, and Aleksander's Recognition Device) [9] and it was initially created to recognize images as a hardware architecture [10]. Although WiSARD has been previously categorized as a supervised learning method, and new developments reveal its usage as unsupervised learning as well [11].

WiSARD organizes its structure in discriminators (or neurons), sets of *X* one-bit word RAMs with *n* inputs. One discriminator often determines just one class of many given as input to the classifier. In the training phase, the classifier receives binary vectors (example instances) and each one depicts an image mapped to a *retina*. A *retina* is a division of the input vector in *X* tuples of *n* bits and is frequently rearranged pseudo-randomly. Each *n*-bit tuple represents an address of RAM of $2^n$ positions, with the *n* tuples mapping *X* memories. Figure 1 illustrates this dynamic and exemplifies how WiSARD operates on the training phase. [11]

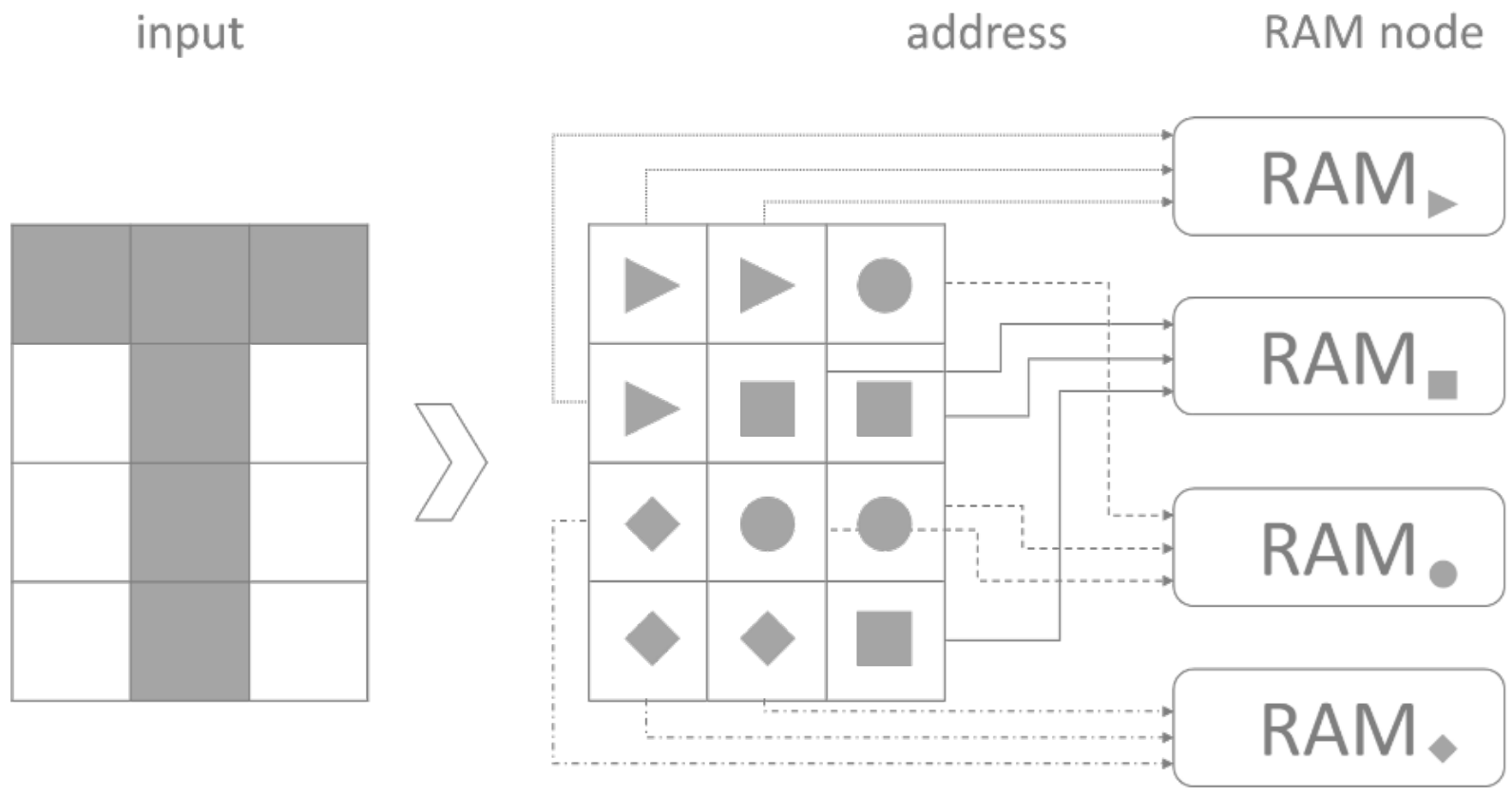

**Fig. 1.** Example of how WiSARD operates on the training phase [11].

The network learns from example writing a ‘1’ in all memory addresses, associated to a particular input. Input mapping in the classification phase follow the same process as in the training examples. After that, all discriminators receive the inputs. The counting of the number of memories mapped by the classification example defines the score $r_y$ for a discriminator $y$. The highest score designates the output class. [11]

The standard WiSARD model may present some drawbacks. One of them occurs when it handles very noisy data. Another problem is saturation, which happens when many classes have a high number of responses from the discriminators. In order to tackle the saturation effect, there is a method called “bleaching” [12]. Sparse retinas can also be a problem, and there is another WiSARD extension to deal with it [13].

## 3 The process dataset

This study analyzed process data from a Brazilian public sector higher education institution specialized in engineering studies. As a public company, it shall keep records of its administrative activities respecting preestablished government rules and laws, which stipulates logged information. As a transaction system, the information system records every time a process moves between organizational units or departments, resulting in a sequence of organizational units visited by each process. Besides, upon creation, information about the kind or processes class, and owner (process metadata) are also recorded.

We dumped the information system database that comprises the logged process data from 2013 to 2017, and it got filtered to remove any inconsistencies. Every resulting process represents a process matrix that describes the process flow through all corresponding organizational units. Figure 2 shows the process matrix where every row represents an organizational unit and the columns represent the sequence in the process life.

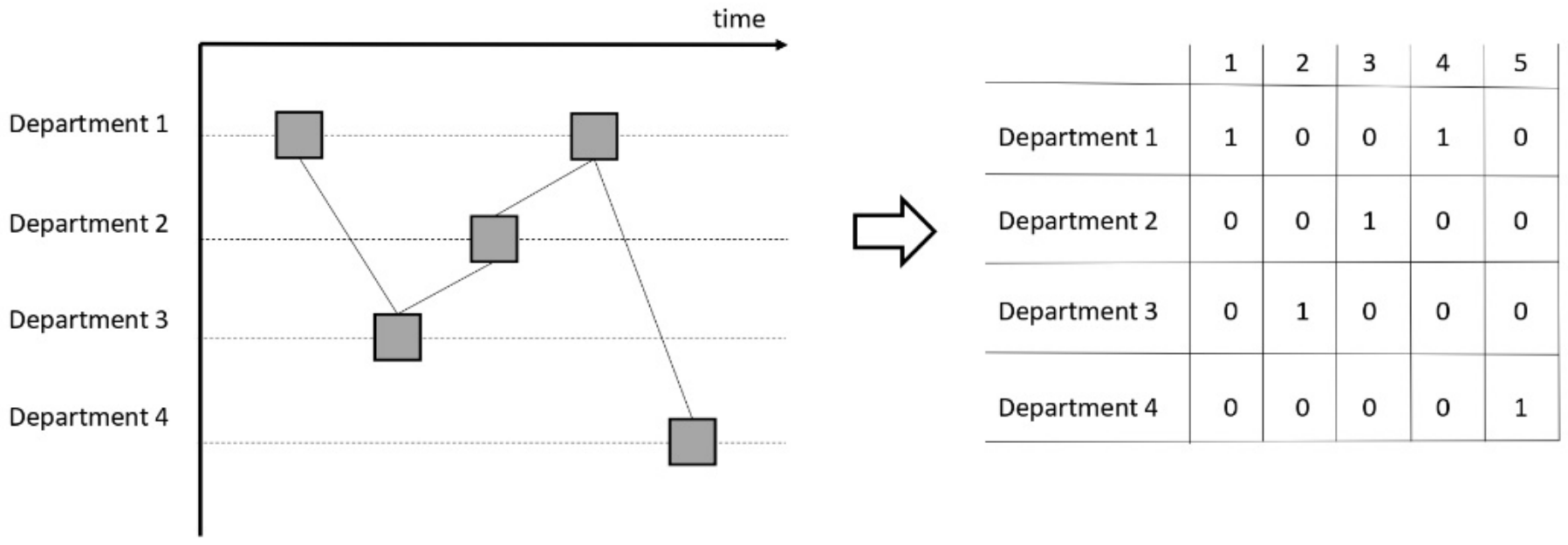

**Fig. 2.** Creation of a retina from process time flow, where each row represents an organizational unit and columns depict the sequence. The final dataset comprises 5,372 administrative processes and each one composed by 173 organizational units versus 78 sequence positions (matrix size of 13,494 elements), and every process was classified accordingly with its kind using the metadata.

The database, summarized in Table 1, exhibits eight different classes and the table displays the process quantity in each class and the amount of unique process matrix or symbols. For example, class A has 1,575 instances, but only 95 distinct ones, showing many repetitions. Conversely, class H shows 467 different processes with 383 distinct symbols. The Shannon entropy [14] and normalized entropy quantify the different inner structure among classes. The both measures are also presented in Table 1, where the Shannon entropy varies from 2.877 to 8.584, and the normalized entropy varies from 0.53434 to 0.98212.

**Table 1**. Dataset summary

| Class | Process set name | Total | Sym | Entropy | Norm. Ent. | Max Pxs | Density |
|---|---|---|---|---|---|---|---|
| A | Internship agreement | 1575 | 95 | 351056 | 0.53434 | 86 | 0.0064 |
| B | Prof. perf assessment | 318 | 25 | 287673 | 0.61947 | 28 | 0.0021 |
| C | Prof. promotion | 349 | 119 | 534360 | 0.77502 | 97 | 0.0072 |
| D | Prof. qualification | 627 | 244 | 676195 | 0.85263 | 135 | 0.0100 |
| E | Staff perf. assessment | 670 | 353 | 732065 | 0.86497 | 214 | 0.0159 |
| F | Procurement | 983 | 696 | 858400 | 0.90901 | 847 | 0.0628 |
| G | Staff promotion | 383 | 245 | 751830 | 0.94729 | 129 | 0.0096 |
| H | Staff qualification | 467 | 383 | 842781 | 0.98212 | 251 | 0.0186 |
| | Total | 5372 | 2160 | - | - | - | - |
| | Min value | 318 | 25 | 2877 | 0.534 | 28 | 0.0021 |
| | Average value | 671.9 | 270 | 6293 | 0.811 | 223.4 | 0.0166 |
| | Max value | 1575 | 696 | 8584 | 0.982 | 847 | 0.0628 |

The lack of proper routine documentation hindered the use of some classes, so we select class A and H in order to determine their process conformance status based on available proper documentation. This subset represents 38% of the initial set and represents the lowest and the highest normalized entropy classes, which provides an idea about the classification performance. Each process on both classes received a tag meaning a conform process (SP - standard process) or non-compliant one (NP - non-conform process).

Classes A and H are depicted on Figure 3, and every square designates a symbol (distinct process matrix), and the number of repetitions appears as a figure in the middle of the representing square. Class A has two important classes where the biggest is an SP type, and the second is an NP. Furthermore, class A has 45% of SP process that are composed by only four symbols. As the entropy shows, class H is more diverse and showcases a greater symbol diversity.

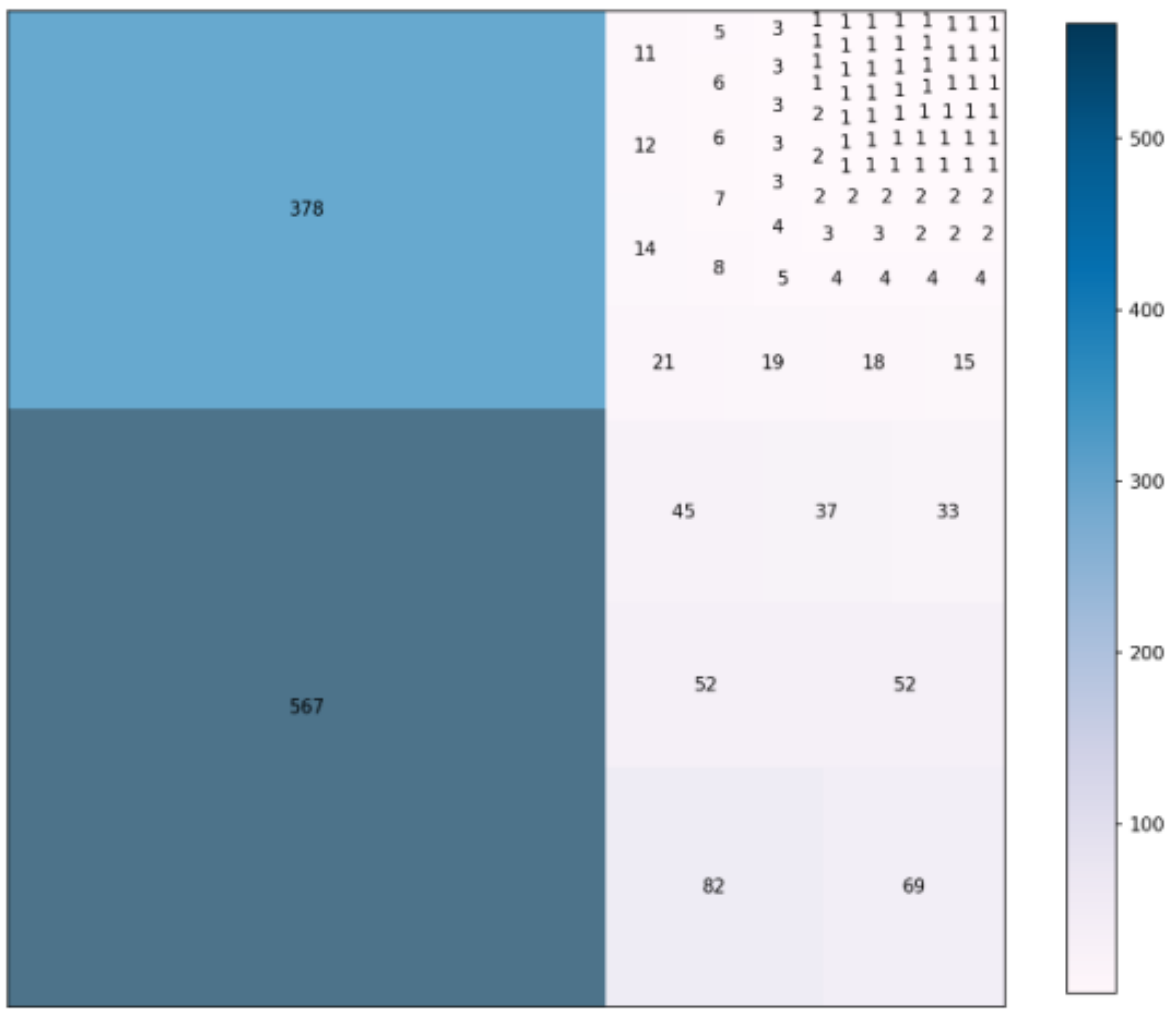

(a) Process class A - Internship agreement

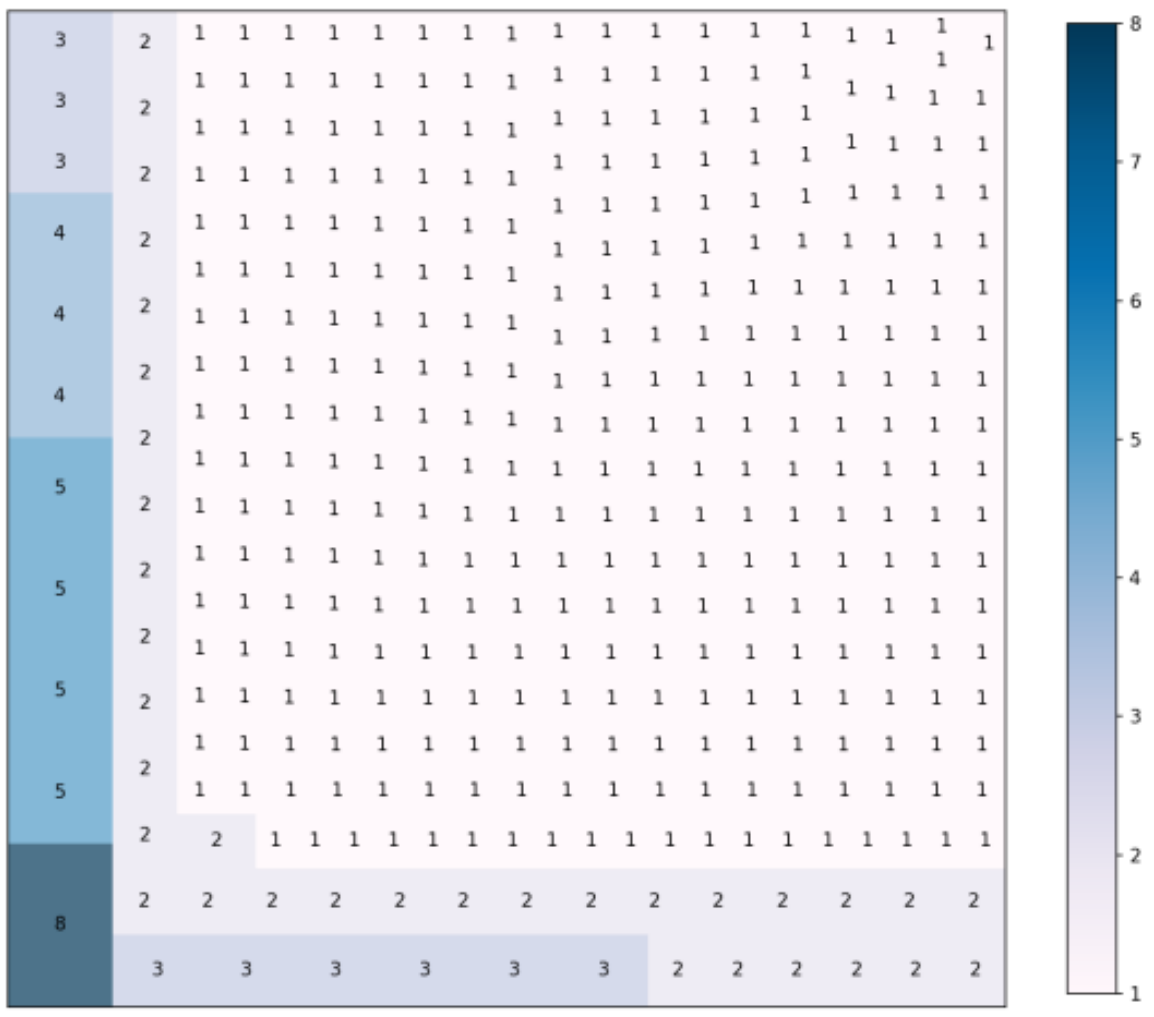

(b) Process class H - Staff qualification

**Fig. 3.** Process classes A (Internship agreement) and H (Staff qualification). Every square designates a symbol (distinct process matrix), and the number of repetitions appears as a figure in the middle of the representing square.

The internship agreement processes (process class A) operationalize Internship Programs for undergraduate students. Since it is mandatory by law, the education institution must examine if enterprises comply with these legal requirements in order to formalize the Agreement. The Agreement allows the enterprise to hire students as interns.

On the other hand, staff qualification processes (process class H) regard the analysis and decision on administrative requests for statutory professional qualification leave petitioned by staff employees.

## 4 The experiment design and results

We compared the WiSARD performance with SVM using 36 LCs, among them, 32 (16 for class A and 16 for class H) were drawn varying the WiSARD's RAM size (2, 4, 8, and 16 bits), the bleaching state (activated or not), and the ignore full-zero patterns (switch on or not). The remaining four curves represent two experiments with SVM for each class, where we tried out two different kernels: a standard linear and a graph kernel (a Weisfeiler-Lehman Graph Kernels) [15]. In addition, we used a public available WiSARD implementation (https://github.com/IAZero/wisardpkg) and SVM from SciKitLearn (https://scikit-learn.org/stable/). Figure 4 shows de resulted curves.

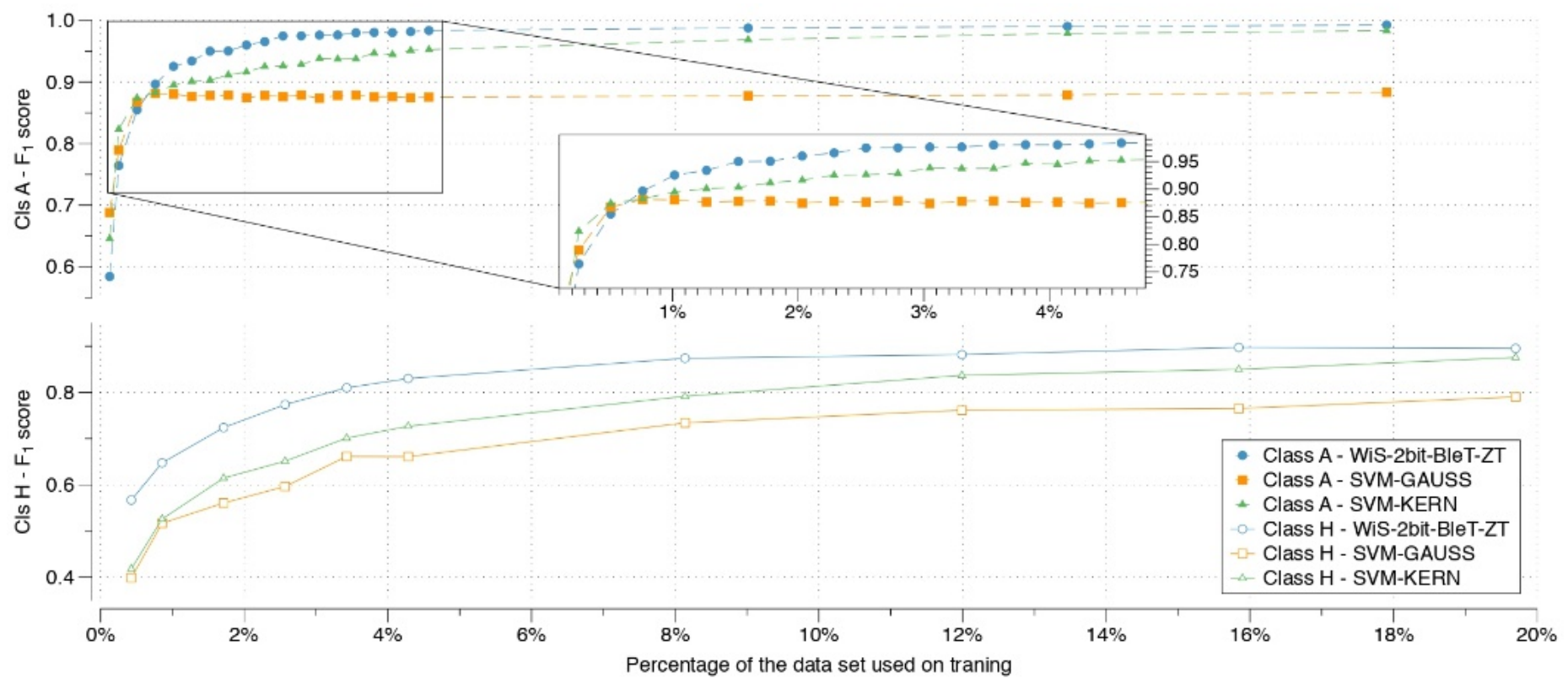

**Fig. 4.** Class A and H $F_1$-Score performance comparison of selected WiSARD configurations (blue lines) with SVM (green and orange lines).

On each LC, we stepped the training size from 2 up to the minimum of the size of SP and NP in each specific class, and then we randomly select each half of the training set respectively from SP and NP subsets. Each point on every LC is the average of the $F_1$-Score of 50 samples.

The comparison of all WiSARD's configurations shows that there is a small improvement with the use of the ignore full-zero pattern flag in this specific context. Although every class has sparse retinas as presented in Table 1, where density varies from 0.0021 to 0.0628 pixels lit per retina on average showing sparseness, improvement is only present within class H (highest density). This effect shows that the lack of bright pixels (meaning that the process has not visited that organizational unit) is so crucial as a lit one. Besides, bleaching increases the $F_1$-Score in 0.0991 on best-case average on every experimented class.

Thus, to choose the best configuration among all 36 WiSARD experiments, we decide to take the configuration: RAM size, bleaching and the ignore full-zero flag that reaches a $F_1$-Score of 0.9 first for each class. Class A got a tie-on RAM size equals 2, 4 or 8, and with bleaching equals true and the ignore full-zero flag as false. This tie happens with a sample size of 12 (6 SP and 6 NC). On the other hand, class H fits better with the WiSARD configured with bleaching, the ignore full-zero flag as true and RAM size of 8 bits at a sample size of 92 (46SP and 46NC).

The only shortcoming presented by the data is that the figure reveals a fluctuation until 0.5% on class A (lower entropy). This abscissa corresponds to a training set composed of 4 SP and 4 NC of 1575 process, and it is due to the unbalanced data in the subset of SP in class A which has cardinality 705, and it comprises only four distinct symbols with the following frequencies: 567, 35, 45, and 52. However, the WiSARD's performance is tantamount with the reference methods.

## 5 Conclusion

In this work, we presented a straightforward graph to retina codification that precludes the kernel construction preprocessing. This representation is more palatable to the process management practitioner, and it avoids any issue related to intermediary representations. Besides, we show that WiSARD generalizes faster and learns with small training sets that help the practitioners since they do not need to further advance on the manual process-dataset classification.